\documentclass{article}

\usepackage{microtype}
\usepackage{graphicx}
\usepackage{subfigure}
\usepackage{booktabs} 
\usepackage{xcolor}
\usepackage{amsmath}
\usepackage{amssymb}
\usepackage{bm}
\usepackage{comment}
\usepackage{caption}
\usepackage{booktabs}

\newcommand{\eg}{\emph{e.g.}}

\DeclareMathOperator{\RePart}{Re}

\newcommand{\envname}{WordCraft} 

\usepackage{hyperref}
\usepackage[capitalise]{cleveref}

\usepackage[accepted]{include/icml2020}
\icmltitlerunning{\envname: An Environment for Benchmarking Commonsense Agents}

\begin{document}

\twocolumn[
\icmltitle{\envname: An Environment for Benchmarking Commonsense Agents}

\icmlsetsymbol{equal}{*}

\begin{icmlauthorlist}
\icmlauthor{Minqi Jiang}{ucl}
\icmlauthor{Jelena Luketina}{ox}
\icmlauthor{Nantas Nardelli}{ox}
\icmlauthor{Pasquale Minervini}{ucl}
\\
\icmlauthor{Philip H. S. Torr}{ox}
\icmlauthor{Shimon Whiteson}{ox}
\icmlauthor{Tim Rockt{\"a}schel}{ucl}
\end{icmlauthorlist}

\icmlaffiliation{ucl}{University College London, London, UK}
\icmlaffiliation{ox}{University of Oxford, Oxford, UK}

\icmlcorrespondingauthor{Minqi Jiang}{minqi.jiang.19@ucl.ac.uk}

\icmlkeywords{Machine Learning, ICML}

\vskip 0.3in
]

\newcommand{\mj}[1]{}
\newcommand{\jl}[1]{}
\newcommand{\pa}[1]{}
\newcommand{\nn}[1]{}
\newcommand{\pt}[1]{}
\newcommand{\sw}[1]{}
\newcommand{\tim}[1]{}
\newcommand{\replace}[2]{}

\printAffiliationsAndNotice{\icmlEqualContribution} 

\newcommand{\acro}{SeRL} 
\newcommand{\longacro}{Semantic Reinforcement Learning} 

\begin{abstract}
    The ability to quickly solve a wide range of real-world tasks requires a commonsense understanding of the world. Yet, how to best extract such knowledge from natural language corpora and integrate it with reinforcement learning (RL) agents remains an open challenge. 
    This is partly due to the lack of lightweight simulation environments that sufficiently reflect the semantics of the real world and provide knowledge sources grounded with respect to observations in an RL environment. To better enable research on  agents making use of commonsense knowledge, we propose \envname, an RL environment based on Little Alchemy 2.
    This lightweight environment is fast to run and built upon entities and relations inspired by real-world semantics. We evaluate several representation learning methods on this new benchmark and propose a new method for integrating knowledge graphs with an RL agent. 
\end{abstract}
\section{Introduction} \label{sec:intro}

Recent progress in Reinforcement Learning (RL), while impressive \citep{silver2016mastering, vinyals2019alphastar, DBLP:journals/corr/abs-1912-06680}, has mostly focused on \emph{tabula-rasa} learning, rather than exploitation of prior world knowledge. 
For example, while Go is an extremely difficult game to master, its rules and thus environment dynamics are simple and not heavily reliant on knowledge about concepts that can be encountered in the real world.
Focusing on tabula-rasa learning confines state-of-the-art approaches to simulation environments that can be solved without the transfer of commonsense, world, or domain knowledge. 
%

%
Many real-world applications (\eg{} personal assistants and household robots) require agents that can learn fast and generalise well to novel situations, which is likely not possible without the ability to reason with commonsense and general knowledge about the world. 
Consider, for example, an agent tasked with performing common household chores that has never seen a dirty ashtray. When presented with this new object, the agent needs to know that a reasonable set of actions involving this object include cleaning the ashtray, but not feeding it to the cat. 
\jl{relating everyday objects? }


\begin{figure}[t]
    \centering
    \includegraphics[width=0.95\columnwidth]{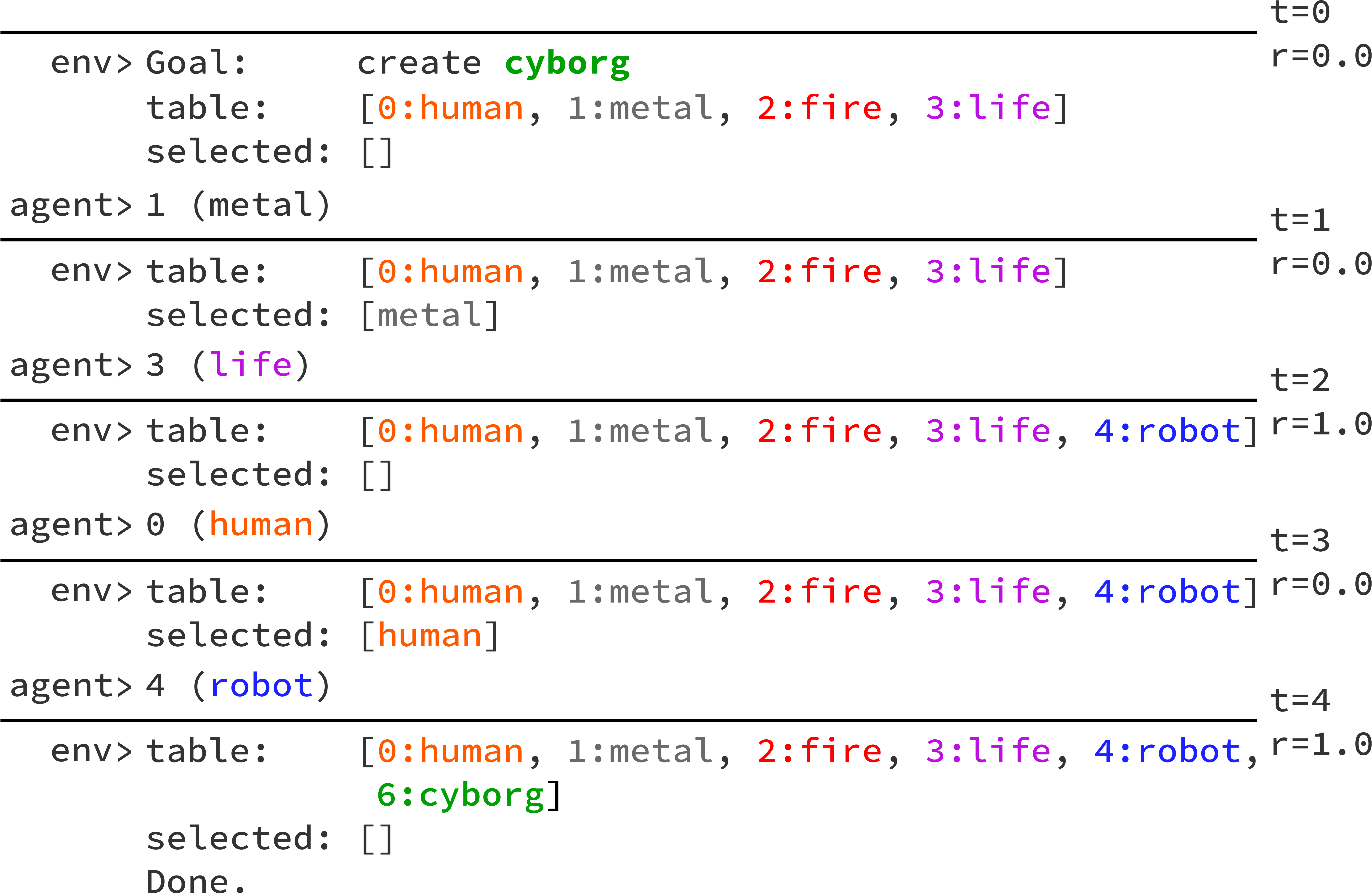}
    \caption{
    Sample episode of \envname: The agent needs to create the goal entity (\emph{cyborg}) from a set of starting entities.}
    \label{fig:env}
\end{figure}

Humans encode a large amount of commonsense knowledge in written language. For example, Wikipedia encodes such knowledge implicitly through corpus statistics, and at times explicitly in writing.
Outside of RL, learning to represent and utilise prior knowledge has improved considerably over recent years.
For example, a common approach is to pre-train neural language models and fine-tune them on downstream tasks~\citep{DBLP:conf/naacl/DevlinCLT19,DBLP:journals/corr/abs-1910-10683}.
Recent works highlight that such pre-trained models capture many aspects of commonsense~\citep{DBLP:journals/corr/abs-1910-01157} and relational knowledge~\citep{DBLP:conf/emnlp/PetroniRRLBWM19}, in addition to linguistic knowledge~\citep{DBLP:conf/acl/JawaharSS19,DBLP:conf/naacl/HewittM19,DBLP:conf/nips/ReifYWVCPK19}.
Another popular approach is to represent commonsense knowledge explicitly as Knowledge Graphs (KGs), as done by ATOMIC~\citep{DBLP:conf/aaai/SapBABLRRSC19} and ConceptNet~\citep{speer2017conceptnet}. Such KGs have been used in commonsense reasoning tasks ~\citep{lin-etal-2019-kagnet}.
%
%

However, the question of how to best utilise commonsense knowledge (provided in language corpora or KGs) for downstream RL tasks has not received the same level of attention, partly due to a scarcity of simulation environments that directly benefit from transferring such prior knowledge. 
%
Robotics environments such as MuJoCo require knowledge not readily expressed in natural language, such as an intuitive understanding of physics \citep{todorov2012mujoco, yu2019meta}.  
%
Rule-based game environments such as Go or Chess \citep{silver2017mastering, silver2018general}, while very hard to master, are based on simple transition dynamics and largely disconnected from entities and concepts encountered in the real world.
On the other hand, commonsense knowledge could be helpful in environments such as StarCraft II  \citep{vinyals2019alphastar, DBLP:journals/corr/abs-1912-06680}, which contain many complex interactions with a large number of entities (many of which have real-world analogues\tim{formal language!}). However, such environments are costly to run, and learning to ground knowledge from external sources remains challenging, as natural language annotations of states and actions in such games are not readily available.
%


To study agents with commonsense knowledge, we present \envname{}, a fast RL environment based on the popular game Little Alchemy 2. 
In this game, the player is tasked with crafting over 700 different entities by combining previously discovered entities. For example, combining ``water'' and ``earth'' creates ``mud''. Learning policies that generalize to unseen entities and combinations requires commonsense knowledge about the world. The environment runs quickly (around 8,000 steps per second on a single machine), enabling fast experimentation cycles. As the entities in the game correspond to words, they can be easily grounded with external knowledge sources. 
In addition, we introduce a method for conditioning agents trained in \envname{} on KGs encoding prior knowledge, such as ConceptNet and other commonsense KGs.
\tim{X\% higher success / more sample efficient training and stronger generalization} \jl{note: we don't test this on commonsense KG, only on ground truth graphs; so maybe "we propose a method for incorporating KGs into RL, that could be used for commonsense KGs?"}

\section{Background}
\label{sec:background}

We formulate the RL problem as a Markov Decision Process (MDP) ~\citep{sutton2018reinforcement} defined by the tuple $(S, A, T, R, \gamma)$, where $s \in S$ is a state, $a \in A$ is an action, $T(s,a) \rightarrow s'$ is the transition function, $R(s, a) \rightarrow \mathbb{R}$ is the reward function, and $\gamma$, the  discount factor. The goal is to find a policy $\pi$ that maximises the expected sum of future discounted rewards: $\sum_{k=0}^{\infty}{\gamma^{k} r_{k + 1}}$.

We represent knowledge about entities and the relations among them as a knowledge graph (KG): a collection of $\langle s, p, o \rangle$ triples, each encoding a relationship of type $p$ (predicate) between the subject $s$ and the object $o$ of the triple.
In order to work with incomplete knowledge graphs and infer missing relationships, we use the ComplEx link prediction  model~\citep{trouillon2016complex} (see \cref{app:background} for details).

\section{\envname}
\label{sec:environment}


In order to benchmark the ability of RL agents to make use of commonsense knowledge, we present \envname, based on Little~Alchemy~2\footnote{\url{https://littlealchemy2.com/}}, a simple association game: Starting from a set of four basic items, the player must create as many different items as possible. Each non-starter item can be created by combining two other items. 
For example, combining ``moon'' and ``butterfly'' yields ``moth'', and combining ``human'' and ``medusa'' yields ``statue''. There are 700 total items (including rare and compound words) and 3,417 permissible item combinations (referred to as valid recipes throughout this paper) \cite{alchemy2wiki}.
Efficiently solving this game without trying every possible combination of items requires using knowledge about relations between common concepts.


\envname~ is a simplified version of Little Alchemy 2, sharing the same entities and valid recipes: First, the interface is text-based rather than graphical. Second, instead of a single open-ended task, \envname~ consists of a large number of simpler tasks. Each task is created by randomly sampling a goal entity, valid constituent entities, and distractor entities. The agent must choose which entities to combine in order to create the goal entity. The task difficulty can be adjusted by increasing the number of distractors or increasing the number of intermediate entities that must be created in order to reach the goal entity (we refer to this variant as \emph{depth-$n$},  with $n-1$ intermediate entities). The training and testing split is created by selecting a set of either goal entities or valid recipes that are not used during training. The relatively large number of entities and valid recipes enables the study of generalization to unseen goals.

The environment is deterministic and fully-observable. At each time step, the state $s$ consists of the goal entity $\mathbf{x}_g$, a set of $k$ currently selectable entities $\mathbf{x}_{\text{table}}=\{\mathbf{x}_i\}_{i=0}^k$, and the currently selected entity $\mathbf{x}_{s,1}$. The action $a$ corresponds to choosing one of the available entities as the second selected entity $\mathbf{x}_{s,2} \in \mathbf{x}_{\text{table}}$. When no entity is currently selected, $\mathbf{x}_{s,1}$ is set to a placeholder empty entity. If $(\mathbf{x}_{s,1}, \mathbf{x}_{s,2})$ corresponds to a valid recipe, the corresponding resulting entity is added to the table $\mathbf{x}_{\text{table}}$ and $\mathbf{x}_{s,1}$ is set to the empty entity. The episode terminates when the newly created entity is $\mathbf{x}_g$ or after a maximum number of steps is reached. We keep the maximum number of steps low to discourage brute-force solutions ($\mathcal{O}(k^2)$ for depth-$1$ tasks). The reward can be sparse (with $r=1$ or $0$ received only at the end of the episode) or shaped (with a penalty for choosing a wrong pair of items or creating irrelevant entities and a reward for creating intermediate ingredients). \jl{think this could go into appendix}

Compared to other environments that combine language and RL \citep{janner_representation_2017, chevalier-boisvert_babyai:_2018, cote_textworld:_2018, das2018embodied}, \envname~ contains a much larger number of objects, with object relationships that reflect the structure of the real world. For example, the underlying KG of TextWorld  (which is perhaps the closest comparison), contains 99 different objects and 10 types of relations \citep{zelinka2019building}. Furthermore, the environment does not contain movement-related actions. The state space is limited to combinations of entity words, and the action space, to table positions, yet generalization remains challenging. This allows allowing us to study commonsense reasoning in isolation, without requiring that the agent learn to ground states to text or deal with a combinatorially large action space (as seen in text games).

\section{Guiding Agents with a Knowledge Graph}
\label{sec:kg}

We propose an agent architecture that makes use of information from an external KG to guide the agent's policy. At a high level, the model consists of a self-attention-based actor-critic network and an external KG link prediction model. 
Given that the recipes in \envname~are based on real-world semantics among common entities, conditioning on commonsense knowledge present in a KG should enable agents to learn more efficiently by constraining their search space to policies biased toward interactions with underlying commonsense semantics. 

\subsection{Self-attention Actor-Critic Network}
\label{sec:saac}
In order to handle a variable number of selectable entities at each time step and ensure selection decisions are invariant to the table position of entities, our policy network makes use of the scaled dot-product attention mechanism introduced in \cite{vaswani_attention_2017}. The self-attention policy network encodes the features $\mathbf{x}_g$ and current selection features $\mathbf{x}_{s,1:2}$ into attention weights over the set of selectable entities. These attention weights $\alpha_i$ act as the logits to the action distribution over the set of selectable entities $e_i$ at each time step: 
\begin{equation*}
\begin{aligned}
\alpha_i & = \frac{1}{\sqrt{d_k}}Q([\mathbf{x}_g; \mathbf{x}_{s,1}; \mathbf{x}_{s,2}])K(\mathbf{x}_{\text{table}})^{\top} \\
\pi&(a_t=e_i|s_t=\mathbf{x}) = \frac{e^{\alpha_i}}{\sum_je^{\alpha_j}}
\end{aligned}
\end{equation*}

, where $\mathbf{x}_{\text{table}}$ are learned representations of selectable entities.
%

In order to predict value estimates, we first apply a linear transform $W$ to $\mathbf{x}_{\text{table}}$ and encode the state as an attention-weighted sum of these linearly transformed entity features. This encoding is passed through a fully-connected layer to yield value estimates $V(\mathbf{x}) = \text{MLP}(\bm{\alpha}^{\top}W(\mathbf{x}_{\text{table}}))$.

\subsection{Link Prediction Model}

We assume access to a link prediction model $\mathbf{L}$ trained on an external knowledge graph that contains information relevant to the RL task. While our model makes use of ComplEx for link prediction \cite{trouillon2016complex}, in general this choice can be substituted by any other link prediction model. We chose ComplEx for its reliable performance in practice. 

The link prediction model is trained on a set of subject-object-predicate relation triplets of the form $(s,p,o)$ to predict higher likelihood scores for true relation triplets. In the context of \envname, each recipe $\{e_1, e_2\} \xrightarrow{} e_3$ reduces to four relevant relation triplets: $(e_1, \texttt{combinesWith}, e_2)$, $(e_2, \texttt{combinesWith}, e_1)$, $(e_1, \texttt{componentOf}, e_3)$, and $(e_2, \texttt{componentOf}, e_2)$. These triplets, when aggregated across all recipes, forms the recipe graph. In our experiments, we train ComplEx on a subgraph of the recipe graph containing all nodes to ensure full entity coverage.

At each time step, we compute relation score vectors $\bm{u}$ and $\bm{v}$ between the goal and selected entities and each $e_i$ on the table and $e_{s,j}$ in the selection: $\bm{u}_i = \mathbf{L}(e_i, \texttt{combinesWith}, e_{s,j})$, $\bm{v}_i = \mathbf{L}(e_i, \texttt{componentOf}, e_g)$.
%
%
These scores are then mixed component-wise with the attention weights, before passing the mixed weights into the softmax policy head, thereby guiding the agent toward relevant predictions from the KG model. Details on learning the mixing coefficients are provided in \cref{app:mixing-coefficients}.

\section{Experiments}
\label{sec:results}

Our experiments focus on zero-shot generalization performance. We split the set of all valid recipes into train (80\%) and test (20\%) sets. Train tasks are generated such that task goals do not involve any recipes in the test set. The model is trained using the TorchBeast implementation of  IMPALA~\citep{espeholt2018impala,torchbeast2019}. 

\subsection{\envname~Benchmarks}\label{no-kg-benchmarks}

To demonstrate how well the task captures real-world relationships between entities, we represent entities as GloVe embeddings \citep{pennington2014glove} when evaluating the model in Section \ref{sec:saac}. We expect GloVe to help generalization if semantically similar entities have similar uses in this environment.
We assess the zero-shot performance of our agent model without a link prediction module 
on depth-1 
tasks with 1 and 8 distractors.

We also collect a human baseline at the same difficulty settings of \envname. This human baseline serves as an estimate of the zero-shot performance that can be achieved using commonsense and general knowledge (see the Appendix \ref{app:human} for the description of human evaluation protocol). 

\begin{figure}[t!]
\includegraphics[width=8cm]{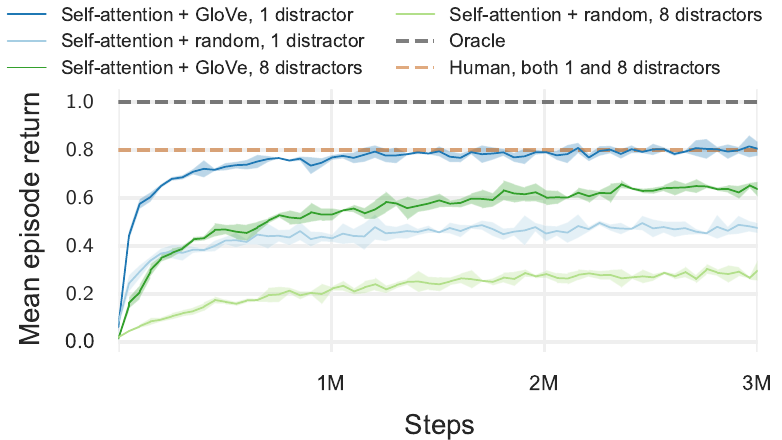}
\caption{Zero-shot success rates over training steps of agents trained using different embeddings on depth-1 tasks with 1 and 8 distractors.}
\label{fig:glove-benchmark-depth1-dis8}
\end{figure}

As seen in \Cref{fig:glove-benchmark-depth1-dis8}, agents trained with GloVe embeddings generalize much better to tasks with recipes that were not part of any training task. This margin of improvement persists in harder task settings, even as the test success rate falls.

\subsection{Knowledge Graph Benchmarks}
We benchmark variants of our full agent with the link prediction module and compare zero-shot performance under different combinations of link prediction scores used to modulate the self-attention weights: both $\texttt{combinesWith}$ and $\texttt{componentOf}$ scores, or each score in isolation. In these experiments, $\mathbf{L}$ is trained on the same train set of recipes used to generate the training tasks.

\begin{figure}[t!]
\includegraphics[width=8cm]{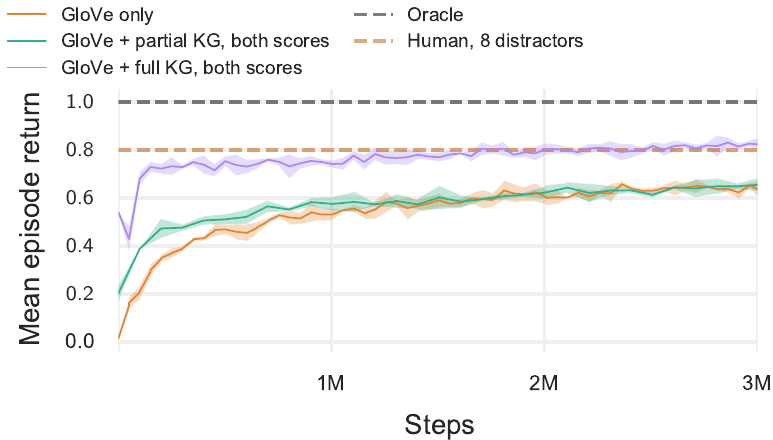}
\caption{Zero-shot success rates over training steps of agents with full and partial KG models on depth-1 tasks with 8 distractors.}
\label{fig:kg-benchmark-depth1-dis8}
\end{figure}

Our results show that access to a full KG model drastically improves the agent's zero-shot performance (see \Cref{fig:kg-benchmark-depth1-dis8}). However, while a partial-KG-model agent reaches an equivalent zero-shot success rate as an agent without any KG model in fewer training steps, they ultimately reach comparable levels of test performance as training progresses (see Appendix for a more detailed discussion).

\section{Related Work} \label{sec:related_work}

Recent work proposes several new environments incorporating a degree of commonsense dynamics and text-base corpora. For example, TextWorld~\citep{cote_textworld:_2018} is a framework that enables creation of novel textual RL environments as well as interfacing with existing text-based games, such as Zork.
\citet{chevalier-boisvert_babyai:_2018, kuettler2020nethack} respectively propose BabyAI and NLE, grid-world environments that evaluate agents on procedurally-generated, goal-conditioned tasks with varying degrees of commonsense structure and available textual corpora.
%
While these environments are challenging in their complex dynamics, large action spaces, sparse rewards, and long planning horizons, these same aspects confound the ability to isolate the problem of conditioning on external knowledge sources.
%
%

\citet{luketina2019survey} presents a comprehensive survey of a related line of research conditioning policies on knowledge in textual corpora or knowledge bases.
\citet{fulda2017can, zahavy2018learn} demonstrate that word embeddings can be an effective modeling tool for policies, while 
\citet{branavan_learning_2012, narasimhan_grounding_2017, zhong2020RTFM} present work that directly conditions policies on domain knowledge encoded in textual corpora.
\citet{marzoev2020unnatural, hill2020human} show that language models can enable generalization to unseen instructions, which indicates that linguistic knowledge can be successfully transferred to aid downstream RL tasks.
Models integrating KGs with agents have been studied in the literature on text-based games \citep{ammanabrolu2019transfer, adhikari2020learning}. Closest to our work, \citet{murugesan2020enhancing} concurrently propose solving a set of TextWorld tasks using a model that incorporates a combination of commonsense knowledge, encoded in ConceptNet, and a learned KG representing the current environment state. They find that while at times beneficial, KGs can also provide too strong a prior.

\section{Future Work}

There are multiple avenues that we plan to further explore. Extending \envname{} to the longer horizon setting of the original Little Alchemy 2, in which the user must discover as many entities as possible, could be an interesting setting to study commonsense-driven exploration. We also plan to introduce additional modalities such as images and control into WordCraft.
%
%
Furthermore, we believe the ideas in this work could benefit more complex RL tasks associated with large corpora of task-specific knowledge, such as NLE. This path of research entails further investigation of methods for automatically constructing knowledge graphs from available corpora as well as agents that retrieve and directly condition on natural language texts in such corpora.

\bibliography{bib/nlp4rl,bib/refs_tim,bib/refs_pasquale}
\bibliographystyle{include/icml2020}

\clearpage

\newpage
\appendix

\section{Background} \label{app:background}

\paragraph{Knowledge Graphs}
%
%
%
%
%
Knowledge graphs naturally represent the ontology of concepts underlying commonsense knowledge about the world. In a knowledge graph, nodes represent concepts, and edges between two nodes represent a relationship between the corresponding concepts.
Let $\mathcal{E}$ and $\mathcal{R}$ denote a set of entities and a set of relation types, respectively.
Formally, a KG $\mathcal{G} \subseteq \mathcal{E} \times \mathcal{R} \times \mathcal{E}$ is a set of subject-predicate-object triples $\langle s, p, o \rangle$, with $s, o \in \mathcal{E}$ and $p \in \mathcal{R}$, where each triple encodes a relationship of type $p$ between the subject $s$ and the object $o$.
There are several examples of large-scale KGs encoding commonsense knowledge, both from academia (such as YAGO~\citep{DBLP:conf/www/SuchanekKW07},  DBpedia~\citep{DBLP:conf/semweb/AuerBKLCI07}) and industry (such as Microsoft's Bing Knowledge Graph and the Google Knowledge Graph~\citep{DBLP:journals/cacm/NoyGJNPT19}). Such large-scale knowledge graphs hold an enormous amount of useful prior knowledge about the world, that should be useful for guiding an RL agent in any environment with real-world semantics. 
Especially relevant to this work are ATOMIC~\citep{DBLP:conf/aaai/SapBABLRRSC19} and ConceptNet~\citep{speer2017conceptnet}, two KGs encoding common-sense knowledge.
\paragraph{Representation Learning of KGs}
An effective way of enabling statistical learning on KGs consists of learning continuous, distributed representations, also referred to as \emph{embeddings}, for all entities in a KG. We refer to \citet{nickel2016review} for a recent survey on this topic.
In this work, we use the ComplEx~\citep{trouillon2016complex} link prediction model, with the loss functions and regularizers proposed by \citet{DBLP:conf/icml/LacroixUO18}.
Given a subject-predicate-object triple $\langle s, p, o \rangle \in \mathcal{E} \times \mathcal{R} \times \mathcal{E}$, ComplEx defines the score $\phi(s, p, o)$ of such a triple as:
\begin{equation} \label{eq:complex}
\phi(s, p, o) = \RePart\left( \langle \mathbf{e}_{s}, \mathbf{e}_{p}, \overline{\mathbf{e}_{o}} \rangle \right),
\end{equation}
\noindent where $\langle {}\cdot{},{}\cdot{},{}\cdot{} \rangle$ denotes the tri-linear dot product, $\mathbf{e}_{s}, \mathbf{e}_{p}, \mathbf{e}_{o} \in \mathbb{C}^{k}$ denote the complex-valued $k$-dimensional representations of $s$, $p$, and $o$, $\overline{\mathbf{x}}$ denote the complex conjugate of $\mathbf{x} \in \mathbb{C}^{k}$, $\RePart\left( \mathbf{x} \right)$ is the real part of $\mathbf{x}$.
In order to learn the embedding representations $\Theta$ for all entities and relation types in a KG $\mathcal{G}$, we follow \citet{DBLP:conf/icml/LacroixUO18} and minimise the following objective via gradient-based optimisation:
\begin{equation} \label{eq:loss}
\begin{aligned}
\mathcal{L}(\mathcal{G}) = & \sum_{\langle s, p, o \rangle \in \mathcal{G}} \left[ \log \sum_{\hat{o} \in \mathcal{E}} \exp \phi(s, p, \hat{o}) \right] \\
 & \quad + \left[ \log \sum_{\hat{s} \in \mathcal{E}} \exp \phi(\hat{s}, p, o) \right] - 2 \phi(s, p, o),
\end{aligned}
\end{equation}
\noindent regularized via the weighted nuclear $p$-norm regularizer proposed by \citet{DBLP:conf/icml/LacroixUO18}.
%


\section{Model Details}

\subsection{Learning Mixing Coefficients} \label{app:mixing-coefficients}

The link-prediction scores between goal and table entities for the $\texttt{componentOf}$ relation and between each selected entity and table entities for the $\texttt{combinesWith}$ relation are combined component-wise with the self-attention weights $\alpha_i$ using mixing coefficients to yield final policy logits $\beta$:

\begin{equation}
    \beta_i = \lambda_\alpha\alpha_i + \lambda_u\bm{u}_i + \lambda_v\bm{v}_i
\end{equation}

\begin{equation}
    \pi(a_t=e_i|s_t=\mathbf{x}) = \frac{e^{\beta_i}}{\sum_je^{\beta_j}}.
\end{equation}

The mixing coefficients $\bm{\lambda}$ are computed by passing the self-attention-weighted table encoding through a fully-connected layer:

\begin{equation}
    \bm{\lambda} = \text{MLP}(\bm{\alpha}^TW_{\lambda}(\mathbf{x}_{table}))
\end{equation}


\section{Experiments}

\subsection{Self-Attention Actor-Critic Architecture}
\begin{figure}[!htbp]
  \centering
  \includegraphics{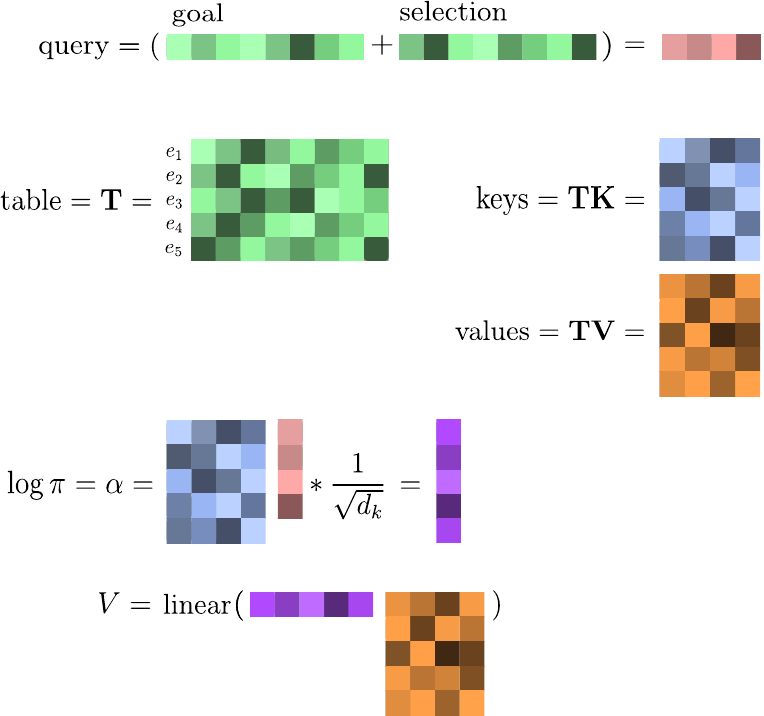}
\caption{Overview of the self-attention actor-critic model used in our experiments.}
\label{fig:selfattn-overview}
\end{figure}

\clearpage

\subsection{Training Parameters}

\begin{table}[!htbp]
\caption{Parameters used for training IMPALA in all experiments.}
\centering
\begin{tabular}{@{}lllll@{}}
\toprule
\textbf{Parameter} & \textbf{Value} \\ \midrule
 Training set ratio & 0.8 \\ 
 Discounting ($\gamma$) & 0.99 \\ 
 Learning rate & 0.001   \\
 Batch size & 128   \\
 Unroll length & 2 \\
 RMSProp $\epsilon$ & 0.01  \\
 Entropy regularization & 0.01   \\
 Reward clipping & none   \\
 Total steps & 3,000,000 \\ \bottomrule
\end{tabular}
\end{table}

\subsection{Model Parameters} 

\begin{table}[!htbp]
\caption{Model parameters used in our experiments.}
\centering
\begin{tabular}{@{}lllll@{}}
\toprule
\textbf{Parameter} & \textbf{Value} \\ \midrule
 Self-attention key size & 300 \\ 
 Self-attention value size & 300 \\ 
 ComplEx embedding size & 128  \\
 Simple-MLP hidden size & 300 \\\bottomrule
\end{tabular}
\end{table}

\subsection{Extended Discussion}

\begin{figure}[!htbp]
  \centering
  \includegraphics{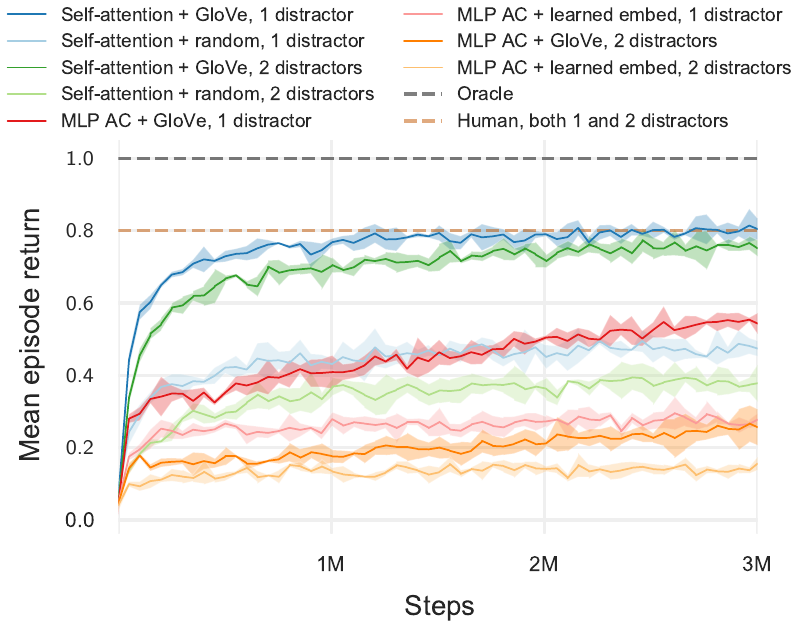}
\caption{Zero-shot success rates of agents trained using different embeddings on depth-1 tasks with 1 and 2 distractors.}
\label{fig:glove-benchmark-depth1-dis2}
\end{figure}

In addition to initial benchmarks conducted in \ref{no-kg-benchmarks}, we also compare the performance of the self-attention actor-critic agent described in \ref{sec:saac} to a simple, single-layer MLP that predicts the policy logits and value estimates from the concatenation of all GloVe word embeddings of goal, selection, and table entities at each time step. These results are presented in \cref{fig:glove-benchmark-depth1-dis2}.

We further benchmark the performance of the self-attention actor-critic agent with and without access to an instance of ComplEx trained on a subgraph of the full recipe graph. We compare agents trained with access to ComplEx trained on a partial graph and those trained on the full graph. The partial KG models are trained on only the relations present in the set of training recipes. 

\begin{figure}[!htbp]
\includegraphics[width=8cm]{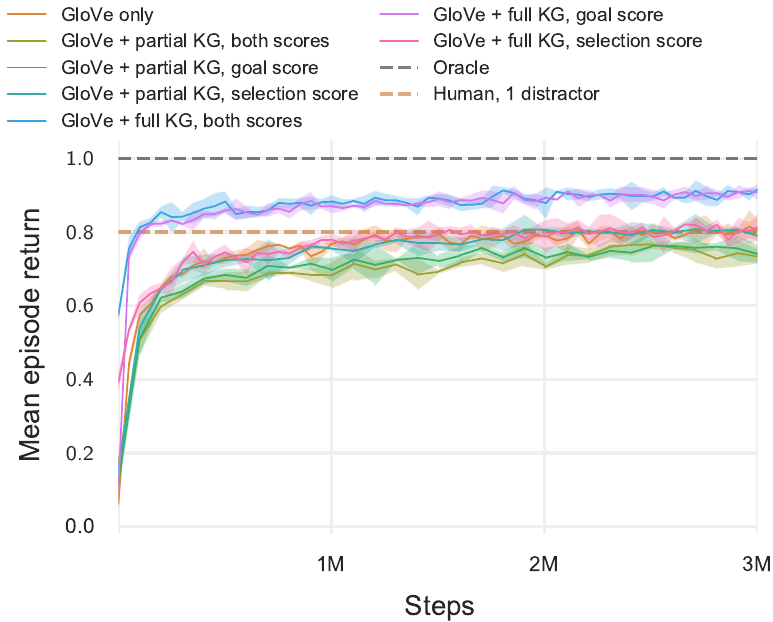}
\caption{Zero-shot success rates of variations of partial- and full-KG agents on depth-1 tasks with 1 distractor. Full KG refers to a ComplEx model trained on the full recipe graph, while partial KG refers to one trained on only training recipes.}
\label{fig:kg-benchmark-depth1-dis1-all}
\end{figure}

\begin{figure}[!htbp]
\includegraphics[width=8cm]{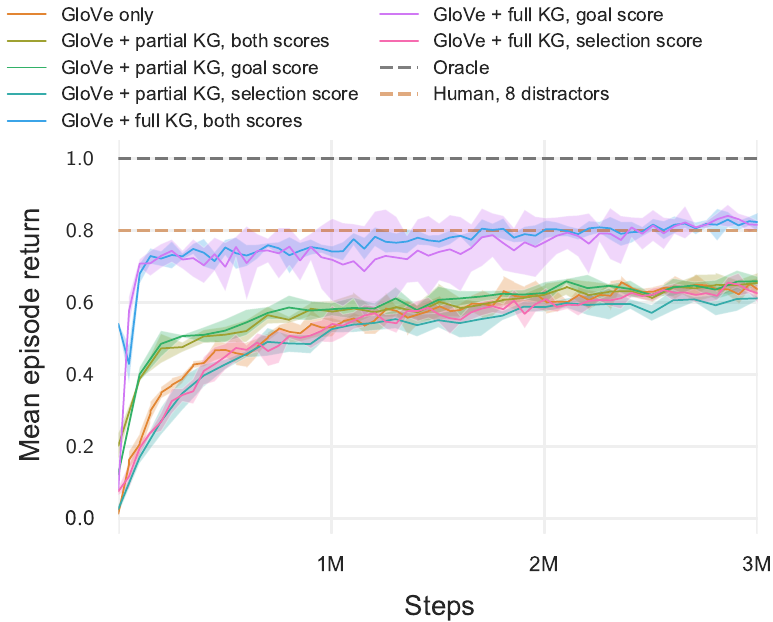}
\caption{Zero-shot success rates of agents trained with and without KG models on depth-1 tasks with 8 distractors.}
\label{fig:kg-benchmark-depth1-dis8-all}
\end{figure}

As the method for incorporating ComplEx's predicted relation scores into the policy prediction performs a weighted sum of each score with the self-attention weights, we investigate the effect of only using different subsets of the scores. As seen in \cref{fig:kg-benchmark-depth1-dis1-all} and \cref{fig:kg-benchmark-depth1-dis8-all}, the subset of relation scores that are helpful to policy learning varies across task depth and distractor settings.

We generally find that access to a full KG model drastically improves the agent's zero-shot success rate, as might be expected. In fact, one might expect that having access to the ground-truth recipe graph as encoded by ComplEx should enable the agent to achieve 100\% zero-shot success rates across all task settings. In order to test the extent to which the full-KG ComplEx scores are predictive of the best actions, we also assessed the performance of agents whose final policy logit mixture only assigned positive weights to different subsets of the relation scores. 
Such full-KG-only agents that choose items only based on the $\texttt{combinesWith}$ relation score will still often choose the same entity twice, believing the entity combines with itself. This is likely because the recipe graph includes several entities that combine with themselves, while the set of all self-combination triplets are not included in the training set. 

Full-KG-only agents that choose items only based on the $\texttt{componentOf}$ relation score will tend to choose the same item twice, as this approach will always assign the highest action probability to selecting the entity on the table that ComplEx predicts to be most likely to be a component of the goal entity.

Interestingly, partial-KG-agents do not always do much better than a KG-free agent. This implies that as the KG-free agent sees more training tasks, it becomes comparable to ComplEx in predicting unseen relations among entities. 

Early on in training, the relation scores predicted by ComplEx seem to help the agent achieve higher zero-shot success rates in fewer training steps compared to KG-free agents (see \Cref{fig:kg-benchmark-depth1-dis8-all}).

\subsection{Human Evaluation Protocol} \label{app:human}

We estimate the generalization ability of humans players on depth-1 tasks with 1 and 8 distractors. Each difficulty setting is tested on a different individual, and we test one individual per setting. The selected individuals were not familiar with the game and got to train on 40 randomly sampled training tasks before being evaluated on 40 testing tasks. The train and test split was the same as in the RL experiments. Surprisingly, we found that changing the number of distractors from 1 to 8 did not significantly influence human performance. We plan to conduct a larger-scale human performance evaluation.

\newpage
\section{Example recipes}

\begin{table}[!htbp]
\caption{Example \envname~ recipes.}
\centering
\begin{tabular}{@{}ll@{}}
\toprule
\textbf{Resulting entity} & \textbf{Entity combination(s)} \\ \midrule
 airplane & bird metal; bird, steel \\
 alcohol & fruit, time; juice, time \\
 batter & flour, milk \\
 cereal & wheat, milk \\ 
 catnip & cat, plant \\ 
 charcoal & fire, wood \\
 dew & water, grass; fog, grass \\
 farmer & human, field; human, plant; \\
 geyser & steam, earth \\
 glacier & ice, mountain \\
 hay bale & hay, hay \\
 iced tea & ice, tea \\
 ivy & plant, wall \\
 juice & water, fruit; pressure, fruit \\
 kite & wind, paper; sky, paper \\
 lake & water, pond; river, dam \\
 milk & farmer, cow; cow, human \\
 milk shake & milk, ice cream \\
 narwhal & unicorn, ocean; unicorn, water \\
 oasis & desert, water \\
 paper & wood, pressure \\
 pasta & flour, egg \\
 rainbow & rain, sun; rain, light \\
 reindeer & Santa, wild animal; livestock, Santa \\
 sand castle & sand, castle \\
 Santa & human, Christmas tree \\
 scythe & blade, grass; blade, wheat \\
 telescope & glass, sky; glass, star; glass, space \\
 umbrella & tool, rain; rain, fabric \\
 volcano & lava, earth; lava, mountain \\
 wallet & leather, money \\
 watch & human, clock \\
 x-ray & light, bone; light, skeleton \\
 yogurt & milk, bacteria \\
 zombie & corpse, life \\
 \bottomrule
\end{tabular}
\end{table}

\end{document}